\documentclass[fleqn,10pt]{wlscirep}
\usepackage[utf8]{inputenc}
\usepackage[T1]{fontenc}
\usepackage{lineno}
\usepackage{multirow}
\usepackage{xcolor,colortbl}
\usepackage{xcolor}
\nolinenumbers
\usepackage{dblfloatfix}

\title{Benchmarking for Biomedical Natural Language Processing Tasks with a Domain Specific ALBERT}

\author[1,*]{Usman Naseem}
\author[2,]{Adam G. Dunn}
\author[1]{Matloob Khushi}
\author[1]{Jinman Kim}
\affil[1]{School of Computer Science, The University of Sydney, Sydney, 2006, Australia}
\affil[2]{School of Medical Science, The University of Sydney, Sydney, 2006, Australia}

\affil[*]{corresponding author: Usman Naseem (usman.naseem@sydney.edu.au)}

\begin{abstract} The availability of biomedical text data and advances in natural language processing (NLP) have made new applications in biomedical NLP possible. Language models trained or fine-tuned using domain-specific corpora can outperform general models, but work to date in biomedical NLP has been limited in terms of corpora and tasks. We present BioALBERT, a domain-specific adaptation of A Lite Bidirectional Encoder Representations from Transformers (ALBERT), trained on biomedical (PubMed and PubMed Central) and clinical (MIMIC-III) corpora and fine-tuned for 6 different tasks across 20 benchmark datasets. Experiments show that BioALBERT outperforms the state-of-the-art on named-entity recognition (+11.09\% BLURB score improvement), relation extraction (+0.80\% BLURB score), sentence similarity (+1.05\% BLURB score), document classification (+0.62\% F1-score), and question answering (+2.83\% BLURB score). It represents a new state-of-the-art in 17 out of 20 benchmark datasets. By making BioALBERT models and data available, our aim is to help the biomedical NLP community avoid computational costs of training and establish a new set of baselines for future efforts across a broad range of biomedical NLP tasks.

\end{abstract}
\begin{document}

\flushbottom
\maketitle

\thispagestyle{empty}


\section*{Background \& Summary}

The growing volume of the published biomedical literature, such as clinical reports~\cite{meystre2008extracting} and health literacy~\cite{maartensson2012health} demands more precise and generalized biomedical natural language processing (BioNLP) tools for information extraction. The recent advancement of using deep learning (DL) in natural language processing (NLP) has fueled the advancements in the development of pre-trained language models (LMs) that can be applied to a range of tasks in the BioNLP domains~\cite{storks2019recent}.


However, directly fine-tuning of the state-of-the-art (SOTA) pre-trained LMs for bioNLP tasks, like Embeddings from Language Models (ELMo)~\cite{peterN18-1202}, Bidirectional Encoder Representations from Transformers (BERT)~\cite{devlin2019bert} and A Lite Bidirectional Encoder Representations from Transformers (ALBERT)~\cite{lan2019albert}, yielded poor performances because these LMs were trained on general domain corpus (e.g. Wikipedia, Bookcorpus etc.), and were not designed for the requirements of biomedical documents that comprise of different word distribution, and having complex relationship~\cite{krallinger2017overview}. To overcome this limitation, BioNLP researchers have trained LMs on biomedical and clinical corpus and proved its effectiveness on various downstream tasks in BioNLP tasks~\cite{Pyysalo2013DistributionalSR, jin2019probing, Si_2019, beltagy2019scibert, peng2019transfer,lee2019biobert,gu2020domain,yuan2021improving}.

Jin et al.~\cite{jin2019probing} trained biomedical ELMo (BioELMo) with PubMed abstracts and found features extracted by BioELMo contained entity-type and relational information relevant to the biomedical corpus. Beltagy et al.~\cite{beltagy2019scibert} trained BERT on scientific texts and published the trained model as Scientific BERT (SciBERT). Similarly, Si et al.~\cite{Si_2019} used task-specific models and enhanced traditional non-contextual and contextual word embedding methods for biomedical named-entity-recognition (NER) by training BERT on clinical notes corpora. Peng et al. ~\cite{peng2019transfer} presented a  BLUE (Biomedical Language Understanding Evaluation) benchmark by designing 5 tasks with 10 datasets for analysing natural biomedical LMs. They also showed that BERT models pre-trained on PubMed abstracts and clinical notes outperformed other models which were trained on general corpora. The most popular biomedical pre-trained LMs is BioBERT (BERT for Biomedical Text Mining)~\cite{lee2019biobert} which was trained on PubMed and PubMed Central (PMC) corpus and fine-tuned on 3 BioNLP tasks including NER, Relation Extraction (RE) and Question Answering (QA). Gu et al.~\cite{gu2020domain} developed PubMedBERT by training from scratch on PubMed articles and showed performance gained over models trained on general corpora. They developed a domain-specific vocabulary from PubMed articles and demonstrated a boost in performance on the domain-specific task. Another biomedical pre-trained LM is KeBioLM~\cite{yuan2021improving} which leveraged knowledge from the UMLS (Unified Medical Language System) bases. KeBioLM was applied to 2 BioNLP tasks. Table~\ref{compdatset} summarises a number of datasets previously used to evaluate Pre-trained LMs on various  BioNLP tasks. Our previous preliminary work has shown the potential of designing a customised domain-specific LM outperforming SOTA in NER tasks~\cite{naseem2020bioalbert}.

\begin{table}[!t]
\centering
\small
\caption{Comparison of the biomedical datasets in prior language model pretraining studies and ours (BioALBERT) -- a biomedical version ALBERT language model }
\label{compdatset}
\begin{tabular}{ccccccc}
\hline \hline
Datasets & BioBERT~\cite{lee2019biobert} & SciBERT~\cite{beltagy2019scibert} & BLUE~\cite{peng2019transfer} & PubMedBERT~\cite{gu2020domain} & KeBioLM~\cite{yuan2021improving} & BioALBERT \\ \hline \hline
Share/Clefe~\cite{suominen2013overview} & $\times$ & $\times$ & \checkmark & $\times$ & $\times$ & \checkmark \\
BC5CDR (Disease)~\cite{Li2016BioCreativeVC} & \checkmark & \checkmark & \checkmark & \checkmark & \checkmark & \checkmark \\
BC5CDR (Chemical)~\cite{Li2016BioCreativeVC} & \checkmark & \checkmark & \checkmark & \checkmark & \checkmark & \checkmark \\
JNLPBA~\cite{10.5555/1567594.1567610} & \checkmark &$\times$  & $\times$ & \checkmark & \checkmark & \checkmark \\
LINNAEUS~\cite{gerner2010linnaeus} & \checkmark & $\times$ & $\times$ & $\times$ & $\times$ & \checkmark \\
NCBI (Disease)~\cite{10.5555/2772763.2772800} & \checkmark & \checkmark &$\times$  & \checkmark & \checkmark & \checkmark \\
Species-800 (S800)~\cite{s800} & \checkmark & $\times$ &$\times$  & $\times$ &$\times$  & \checkmark \\
BC2GM~\cite{Ando2007BioCreativeIG} & \checkmark & $\times$ & $\times$ & \checkmark & \checkmark & \checkmark \\ \hline \hline
DDI~\cite{herrero2013ddi} & $\times$ &  $\times$& \checkmark & \checkmark & \checkmark & \checkmark \\
ChemProt~\cite{krallinger2017overview} & \checkmark & \checkmark & \checkmark & \checkmark & \checkmark & \checkmark \\
i2b2~\cite{uzuner20112010} &$\times$  & $\times$ & \checkmark & $\times$ & $\times$ & \checkmark \\
Euadr~\cite{van2012eu} & \checkmark & $\times$ & $\times$ &$\times$  & $\times$ & \checkmark \\
GAD~\cite{bravo2015extraction} & \checkmark & $\times$ & $\times$ & \checkmark & \checkmark & \checkmark \\ \hline \hline
BIOSSES~\cite{souganciouglu2017biosses}  &$\times$  &$\times$  & \checkmark & \checkmark & $\times$ & \checkmark \\
MedSTS~\cite{wang2020medsts} & $\times$ & $\times$ & \checkmark & $\times$ & $\times$ & \checkmark \\ \hline \hline
MedNLI~\cite{romanov2018lessons} & $\times$ & $\times$ & \checkmark & $\times$ & $\times$ & \checkmark \\ \hline \hline
HoC~\cite{baker2016automatic} &$\times$  &$\times$ & \checkmark & \checkmark & $\times$ & \checkmark \\ \hline \hline
BioASQ 4b~\cite{tsatsaronis2015overview} & \checkmark &$\times$  & $\times$ & \checkmark & $\times$ & \checkmark \\
BioASQ 5b~\cite{tsatsaronis2015overview} & \checkmark & $\times$ &$\times$  & \checkmark & $\times$ & \checkmark \\
BioASQ 6b~\cite{tsatsaronis2015overview} & \checkmark & $\times$ &$\times$  & \checkmark & $\times$ & \checkmark \\ \hline \hline
\end{tabular}
\end{table}








With all these pre-trained LMs adopting BERT architecture, its’ training is slow and requires huge computational resources. Further, all these LMs were demonstrated with selected BioNLP tasks, and therefore their generalizability is unproven. Furthermore, these LMs are trained on limited domain-specific corpora, whereas some tasks contain both clinical and biomedical terms, so training with broader coverage of domain-specific corpora can improve performance. ALBERT has been shown to be a superior model compared to BERT in NLP tasks~\cite{lan2019albert}, and we suggest that this model can be trained to improve BioNLP tasks as shown with BERT. In this study, we hypothesize that training ALBERT on biomedical (PubMed and PMC) and clinical notes (MIMIC-III) corpora can be more effective and computationally efficient in BioNLP tasks as compared to other SOTA methods.





We present biomedical ALBERT (BioALBERT),  a new LM designed and optimized to benchmark performance on a range of BioNLP tasks. BioALBERT is based on ALBERT and trained on a large corpus of biomedical and clinical texts.  We fined-tuned and compared the performance of BioALBERT on 6 BioNLP tasks with 20 biomedical and clinical benchmark datasets with different sizes and complexity. Compared with most existing BioNLP LMs that are mainly focused on limited tasks, our BioALBERT achieved SOTA performance on 5 out of 6 BioNLP tasks in 17 out of 20 tested datasets. BioALBERT achieved higher performance in NER, RE, Sentence similarity, Document classification and a higher Accuracy (lenient) score in QA  than the current SOTA LMs. To facilitate developments in the important BioNLP community, we make the pre-trained BioALBERT LMs and the source code for fine-tuning BioALBERT publicly available\footnote{https://github.com/usmaann/BioALBERT}.

\section*{Methods} \label{aaamodel}

\begin{figure}[!htpb]
\centering
\includegraphics[width=1\linewidth]{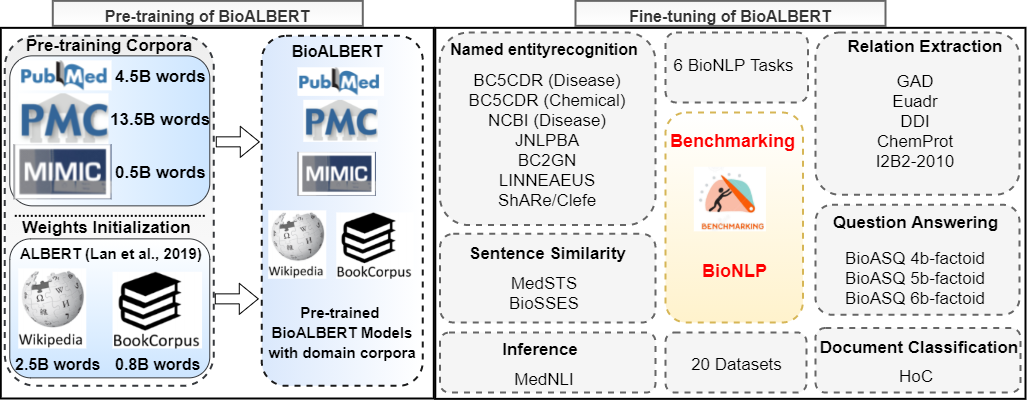}
\caption{An overview of  pre-training, fine-tuning and the diverse tasks and datasets present in Benchmarking for BioNLP using BioALBERT}
\label{ben}
\end{figure}

BioALBERT has the same architecture as ALBERT. The overview of pre-training, fine-tuning, variants of tasks and datasets used for BioNLP is shown in Figure~\ref{ben}. We describe ALBERT and then the pre-training and fine-tuning process employed in BioALBERT.


\subsection*{ALBERT}

ALBERT~\cite{lan2019albert} is built on the architecture of BERT to mitigate a large number of parameters in BERT, which causes model degradation, memory issues and degraded pre-training time. ALBERT is a contextualized LM that is based on a masked language model (MLM) and pre-trained using bidirectional transformers~\cite{vaswani2017attention} like BERT. ALBERT uses an MLM that predicts randomly masked words in a sequence and can be used to learn bidirectional representations. 

ALBERT is trained on the same English Wikipedia and BooksCorpus as in BERT; however, reduced BERT parameters by 87 percent and could be trained nearly twice as fast. ALBERT reduced parameters requirement by factorizing and decomposing large vocabulary embedding matrix into two smaller matrixes. ALBERT’s other improvements are introducing sentence order prediction (SOP) loss instead of next sentence prediction (NSP) and the introduction of cross-layer parameter sharing, which prevents parameters from growing with the depth of the network. In the following section, we describe the steps involved in training BioALBERT.

\subsection*{\textbf{Pre-training BioALBERT}}

We first initialized BioALBERT with weights from ALBERT during the training phase. Biomedical terminologies have terms that could mean different things depending upon its context of appearance. For example, ER could be referred to ‘estrogen receptor’ gene or its product as protein. Similarly, RA may represent ‘right atrium’ or ‘rheumatoid arthritis’ depending upon the context of appearance. On the other hand, two terminologies could be used to refer to a similar concept, such as ‘heart attack’ or ‘myocardial infarction’. As a result, pre-trained LM trained on general corpus often obtains poor performance.
\begin{table}[!htpb]
\centering
\small
\caption{List of text corpora used for BioALBERT}
\label{corpuss}
\begin{tabular}{ccc}
\hline \hline
Corpus & Number of Words & Domain \\ \hline \hline
English   Wikipedia & 2.5 Billion & General \\ \hline
BooksCorpus & 0.8 Billion & General \\ \hline
PubMed   Abstracts & 4.5 Billion & Biomedical \\ \hline
PMC Full-text   articles & 13.5 Billion & Biomedical \\ \hline
MIMMIC-III & 0.5 Billion & Clinical \\ \hline \hline
\end{tabular}
\end{table}


\begin{table}[!t]
\centering
\small
\caption{Summary of parameters used in the Pre-training of BioALBERT}
\label{parameters}
\begin{tabular}{cc}
\hline \hline
\multicolumn{2}{c}{Summary of All parameters used: (Pre-Training)} \\ \hline \hline
Architecture & ALBERT \\ \hline
Activation Function & GeLU \\ \hline
Attention Heads & 12 \\ \hline
No. of Layers & 12 \\ \hline
Size of Hidden Layer & 768 \\ \hline
Size of Embedding & 128 \\ \hline
Size of Vocabulary & 30k \\ \hline
Optimizer Used & LAMB \\ \hline
Training Batch   Size & 1024 \\ \hline
Evaluation Batch   Size & 16 \\ \hline
Maximum Sentence Length & 512 \\ \hline
Maximum Predictions per Sentence & 20 \\ \hline
Warm-up   Steps & 3,125 \\ \hline \hline
\end{tabular}
\end{table}

BioALBERT is the first domain-specific LM trained on biomedical domain corpus and clinical notes. The text corpora used for pre-training of BioALBERT are listed in Table~\ref{corpuss}. BioALBERT is trained on abstracts from PubMed, full-text articles of PMC and clinical notes (MIMIC) and their combination. These unstructured and raw corpora were converted to structured format by converting raw text files into a single sentence in which: (i) within a text, all blank lines were removed and transformed into a single paragraph, and (ii) any line with a length of less than 20 characters was excluded. Overall, PubMed contained approximately 4.5 billion words, PMC contains about 13.5 billion words, and MIMIC contained 0.5 billion words.

We used sentence embeddings for tokenization of BioALBERT by pre-processing the data as a sentence text. Each line was considered as a sentence keeping the maximum length to 512 words by trimming. If the sentence was shorter than 512 words, then more words were embedded from the next line. An empty line was used to define a new document. 3,125 warm-up steps are used for training all of our models. We used LAMB optimizer during the training process of our models and kept the vocabulary size to 30K. GeLU activation is used in all variants of models during the training process. For BioALBERT base models, a training batch size of 1,024 was used, whereas, in BioALBERT large models, training size was reduced to 256 due to computational resources limitations. Summary of parameters used in the training process is given in Table~\ref{parameters}.

We present 8 models (Table~\ref{versions}) consisting of 4 base and 4 large LMs. We identified that on V3-8 TPU,  both base and large LMs were successful with a larger batch size during training. The base model contained 128 embedding size and 12 million parameters, whereas the large model had 256 embedding size and 16 million parameters.

\begin{table}[!htpb]
\centering
\small
\caption{BioALBERT trained on different training steps, different combinations of the text corpora given in Table~\ref{corpuss}, and BioALBERT model version and size}
\label{versions}
\begin{tabular}{cccc}
\hline \hline
\begin{tabular}[c]{@{}c@{}}Model \\   Version.\end{tabular} & BioALBERT Size & \begin{tabular}[c]{@{}c@{}}Combination of corpus\\  used  for training\end{tabular} & \begin{tabular}[c]{@{}c@{}}Number of\\ training steps\end{tabular} \\ \hline \hline
1 & Base1 & Wikipedia + BooksCorpus + PubMed & 200K \\ \hline
2 & Base2 & Wikipedia + BooksCorpus + PubMed+ PMC & 470K \\ \hline
3 & Large1 & Wikipedia + BooksCorpus + PubMed & 200K \\ \hline
4 & Large2 & Wikipedia + BooksCorpusPubMed + PMC & 470K \\ \hline
5 & Base3 & Wikipedia + BooksCorpus + PubMed + MIMIC-III & 200K \\ \hline
6 & Base4 & Wikipedia + BooksCorpus + PubMed + PMC + MIMIC-III & 200K \\ \hline
7 & Large3 & Wikipedia + BooksCorpus + PubMed + MIMIC-III & 270K \\ \hline
8 & Large4 & Wikipedia + BooksCorpus + PubMed + PMC + MIMIC-III & 270K \\ \hline \hline
\end{tabular}
\end{table}

\subsection*{\textbf{Fine-tuning BioALBERT}}
Similar to other SOTA biomedical LMs~\footnote{We followed the same architectural modification as previous studies in the downstream task.}, BioALBERT was tested on a number of downstream BioNLP tasks which required minimal architecture alteration. BioALBERT’s computational requirements were not significantly large compared to other baseline models, and fine-tuning only required relatively small computation compared to the pre-training. BioALBERT employed less physical memory and improvised on parameter sharing techniques, and learned the word embeddings using the sentence piece tokenization, which gives it better performance and faster training compared to other SOTA biomedical LMs.




\begin{table}[!htpb]
\centering
\small
\caption{Summary of parameters used in fine-tuning}
\label{finetuning}
\begin{tabular}{cc}
\hline \hline
\multicolumn{2}{c}{Summary   of All parameters used: (Fine-Tuning)} \\ \hline \hline
Optimizer used & AdamW \\ \hline
Training Batch Size   & 32 \\ \hline
Checkpoint Saved & 500 \\ \hline
Learning Rate & 0.00001 \\ \hline
Training Steps & 10k \\ \hline
Warm-up   Steps & 320 \\ \hline \hline
\end{tabular}
\end{table}

During fine-tuning, we use the weights of the pre-trained BioALBERT LM. We used an AdamW optimizer and a learning rate of 0.00001. A batch size of 32 was used during training. We restricted sentence length to 512 for NER and 128 for all other tasks and lower-cased all words. Finally, we fine-tuned our pre-trained models for 10k training steps and used 512 warm-up steps during the fine-tuning process.  The test datasets were used for prediction, and the evaluation metric was compared with previous SOTA models. Table~\ref{finetuning} summaries all fine-tuning parameters.

\subsection*{\textbf{Tasks and Datasets}}

We fine-tuned BioALBERT on 6 different BioNLP tasks with 20 datasets that cover a broad range of data quantities and difficulties (\textbf{Table~\ref{stats}}). We rely on pre-existing datasets that are widely supported in the BioNLP community and describe each of these tasks and datasets.



\begin{table}[!htpt]
\centering
\small
\caption{Statistics of the datasets used}
\label{stats}
\begin{tabular}{ccccccc}
\hline \hline
Dataset & Task & Domain & Train & Dev & Test & Metric \\ \hline \hline
BC5CDR (Disease) & NER & Biomedical & 109853 & 121971 & 129472 & F1-Score \\
BC5CDR (Chemical) & NER & Biomedical & 109853 & 117391 & 124676 & F1-Score \\
NCBI (Disease) & NER & Clinical & 135615 & 23959 & 24488 & F1-Score \\
JNLPBA & NER & Biomedical & 443653 & 117213 & 114709 & F1-Score \\
BC2GM & NER & Biomedical & 333920 & 70937 & 118189 & F1-Score \\
LINNAEUS & NER & Biomedical & 267500 & 87991 & 134622 & F1-Score \\
Species-800 (S800) & NER & Biomedical & 147269 & 22217 & 42287 & F1-Score \\
Share/Clefe & NER & Clinical & 4628 & 1075 & 5195 & F1-Score \\ \hline \hline
GAD & RE & Biomedical & 3277 & 1025 & 820 &  F1-Score \\
Euadr & RE & Biomedical & 227 & 71 & 57 &  F1-Score \\
DDI & RE & Biomedical & 2937 & 1004 & 979 &  F1-Score \\
ChemProt & RE & Biomedical & 4154 & 2416 & 3458 &  F1-Score \\
i2b2 & RE & Clinical & 3110 & 11 & 6293 &  F1-Score \\ \hline \hline
HoC & Document Classification & Biomedical & 1108 & 157 & 315 & F1-Score \\ \hline \hline
MedNLI & Inference & Clinical & 11232 & 1395 & 1422 & Accuracy \\ \hline \hline
MedSTS & Sentence Similarity & Clinical & 675 & 75 & 318 & Pearson \\
BIOSSES & Sentence Similarity & Biomedical & 64 & 16 & 20 & Pearson \\ \hline \hline
BioASQ 4b-factoid & QA & Biomedical & 327 & -- & 161 & \begin{tabular}[c]{@{}c@{}}Accuracy \\ (Lenient)\end{tabular} \\
BioASQ 5b-factoid & QA & Biomedical & 486 & -- & 150 & \begin{tabular}[c]{@{}c@{}}Accuracy \\ (Lenient)\end{tabular} \\
BioASQ 6b-factoid & QA & Biomedical & 618 & -- & 161 & \begin{tabular}[c]{@{}c@{}}Accuracy \\ (Lenient)\end{tabular} \\ \hline \hline
\end{tabular}
\end{table}

\begin{itemize}
    \item \textbf{Named entity recognition (NER):} Recognition of proper domain-specific nouns in a biomedical corpus is the most basic and important BioNLP task.  F1-score was used as an evaluation metric for NER. BioALBERT was evaluated on 8 NER benchmark datasets (From Biomedical and Clinical domain): We used NCBI (Disease)~\cite{10.5555/2772763.2772800},  BC5CDR (Disease)~\cite{Li2016BioCreativeVC}, BC5CDR (Chemical)~\cite{Li2016BioCreativeVC}, BC2GM~\cite{Ando2007BioCreativeIG}, JNLPBA~\cite{10.5555/1567594.1567610}, LINNAEUS~\cite{gerner2010linnaeus}, Species-800 (S800)~\cite{s800} and Share/Clefe~\cite{suominen2013overview} datasets. 
     
     \item \textbf{Relation Extraction (RE):} RE tasks aim to identify relationship among entities in a sentence. The annotated data were compared with relationship and types of entities. Micro-average F1-score metric was used as an evaluation metric. For RE, we used DDI~\cite{herrero2013ddi}, Euadr~\cite{van2012eu}, GAD~\cite{bravo2015extraction}, ChemProt~\cite{krallinger2017overview} and i2b2~\cite{uzuner20112010} datasets.

     \item \textbf{Document Classification:} Document classification tasks classifies the whole document into various categories. Multiple labels from texts are predicted in the multi-label classification task. For the document classification task, we followed the common practice and reported the F1-score. For document classification, we used HoC (the hallmarks of Cancers)~\cite{baker2016automatic} dataset.

     \item \textbf{Inference:} Inference tasks predict whether the premise sentence entails the hypothesis sentence. It mainly focuses on causation relationships between sentences. For evaluation, we used overall standard accuracy as a metric. For inference, we used MedNLI~\cite{romanov2018lessons} dataset.
     

\item \textbf{Sentence Similarity (STS):} STS task is to predict similarity scores by estimating whether two sentences deliver
similar contents. Following common practise, we evaluate similarity by using
Pearson correlation coefficients. We used MedSTS~\cite{wang2020medsts} and BIOSSES~\cite{souganciouglu2017biosses} datasets for sentence similarity task.

\item \textbf{Question Answering (QA):} QA is a task of answering questions posed in the natural language given related passages.  We used accuracy as an evaluation metric for the QA task. For QA, we used BioASQ factiod~\cite{tsatsaronis2015overview} datasets.

\end{itemize}
\begin{table*}[!t]
\centering
\small
\caption{Comparison of BioALBERT v/s SOTA methods in BioNLP tasks }
\label{results}
\begin{tabular}{ccccccccccc}
\hline \hline
\multirow{2}{*}{Dataset} & \multirow{2}{*}{SOTA} & \multicolumn{8}{c}{BioALBERT} & \multirow{2}{*}{\begin{tabular}[c]{@{}c@{}} Difference over\\  SOTA\end{tabular}} \\ \cline{3-10}
 &  & \begin{tabular}[c]{@{}c@{}}Base1\end{tabular} & \begin{tabular}[c]{@{}c@{}}Base2\end{tabular} & \begin{tabular}[c]{@{}c@{}}Large1\end{tabular} & \begin{tabular}[c]{@{}c@{}}Large2\end{tabular} & \begin{tabular}[c]{@{}c@{}}Base3\end{tabular} & \begin{tabular}[c]{@{}c@{}}Base4\end{tabular} & \begin{tabular}[c]{@{}c@{}}Large3\end{tabular} & \begin{tabular}[c]{@{}c@{}}Large4\end{tabular} &  \\ \hline \hline
\multicolumn{11}{c}{Named Entity Recognition Task} \\ \hline \hline
Share/Clefe & 75.40 & 94.27 & 94.47 & 93.16 & 94.30 & \textbf{94.84} & 94.82 & 94.70 & 94.66 & 19.44 $\uparrow$  \\
\begin{tabular}[c]{@{}c@{}}BC5CDR\\ (Disease)\end{tabular} & 87.15 & 97.66 & 97.62 & \textbf{97.78} & 97.61 & 90.03 & 90.01 & 90.29 & 91.44 & 10.63 $\uparrow$  \\
\begin{tabular}[c]{@{}c@{}}BC5CDR\\ (Chemical)\end{tabular} & 93.47 & 97.90 & \textbf{98.08} & 97.76 & 97.79 & 89.83 & 90.08 & 90.01 & 91.48 & 4.61 $\uparrow$  \\
JNLPBA & 82.00 & 82.72 & 83.22 & 84.01 & 83.53 & \textbf{86.74} & 86.56 & 86.20 & 85.72 & 4.74 $\uparrow$  \\
Linnaeus & 93.54 & 99.71 & 99.72 & 99.73 & \textbf{99.73} & 95.72 & 98.27 & 98.24 & 98.23 & 6.19 $\uparrow$  \\
\begin{tabular}[c]{@{}c@{}}NCBI\\ (Disease)\end{tabular} & 89.71 & 95.89 & 95.61 & \textbf{97.18} & 95.85 & 85.82 & 85.93 & 85.86 & 85.83 & 7.47 $\uparrow$  \\
S800 & 75.31 & 98.76 & 98.49 & \textbf{99.02} & 98.72 & 93.53 & 93.63 & 93.63 & 93.63 & 23.71 $\uparrow$  \\
BC2GM & 85.10 & 96.34 & 96.02 & \textbf{96.97} & 96.33 & 83.35 & 83.38 & 83.44 & 84.72 & 11.87 $\uparrow$  \\ \hline \hline
\textbf{BLURB} & 84.61 & 95.41 & 95.41 & \textbf{95.70} & 95.48 & 89.98 & 90.34 & 90.30 & 90.71& 11.09$\uparrow$  \\ \hline \hline
\multicolumn{11}{c}{Relation Extraction Task} \\ \hline \hline
DDI & 82.36 & 82.32 & 79.98 & 83.76 & \textbf{84.05} & 76.22 & 75.57 & 76.28 & 76.46 & 1.69 $\uparrow$  \\
ChemProt & 77.50 & \textbf{78.32} & 76.42 & 77.77 & 77.97 & 62.85 & 62.34 & 61.69 & 57.46 & 0.82 $\uparrow$  \\
i2b2 & 76.40 & 76.49 & 76.54 & \textbf{76.86} & 76.81 & 73.83 & 73.08 & 72.19 & 75.09 & 0.46 $\uparrow$  \\
Euadr & \textbf{86.51} & 82.32 & 74.07 & 84.56 & 81.32 & 62.52 & 76.93 & 70.41 & 70.48 & -1.95 $\downarrow$ \\
GAD & \textbf{84.30} & 73.82 & 66.32 & 76.74 & 69.65 & 72.68 & 69.14 & 71.81 & 68.17 & -7.56 $\downarrow$\\ \hline \hline
\textbf{BLURB} & 79.14 & 78.66 & 74.67 & \textbf{79.94} & 77.96 & 69.62 & 71.41 & 70.50 & 69.53&0.80$\uparrow$  \\ \hline \hline
\multicolumn{11}{c}{Sentence Similarity Task} \\ \hline \hline
BIOSSES & 92.30 & 82.27 & 73.14 & \textbf{92.80} & 81.90 & 24.94 & 55.80 & 47.86 & 30.48 & 0.50 $\uparrow$  \\
MedSTS & 84.80 & 85.70 & 85.00 & \textbf{85.70} & 85.40 & 51.80 & 56.70 & 45.80 & 42.00 & 0.90 $\uparrow$  \\ \hline \hline
\textbf{BLURB} & 88.20 & 83.99 & 79.07 & \textbf{89.25} & 83.65 & 38.37 & 56.25 & 46.83 & 36.24&1.05$\uparrow$   \\ \hline \hline
\multicolumn{11}{c}{Inference Task} \\ \hline \hline
MedNLI & \textbf{84.00} & 77.69 & 76.35 & 79.38 & 79.52 & 78.25 & 77.20 & 76.34 & 75.51 & -4.48 $\downarrow$ \\ \hline \hline
\multicolumn{11}{c}{Document Classification Task} \\ \hline \hline
HoC & 87.30 & 83.21 & 84.52 & \textbf{87.92} & 84.32 & 64.20 & 75.20 & 61.00 & 81.70 & 0.62 $\uparrow$  \\ \hline \hline
\multicolumn{11}{c}{Question Answering Task} \\ \hline \hline
BioASQ 4b & 47.82 & 47.90 & 48.34 & \textbf{48.90} & 48.25 & 47.10 & 47.35 & 45.90 & 46.10 & 1.08 $\uparrow$  \\
BioASQ 5b & 60.00 & 61.10 & 61.90 & \textbf{62.31} & 61.57 & 58.54 & 59.21 & 58.98 & 58.50 & 2.31 $\uparrow$  \\
BioASQ 6b & 57.77 & 59.80 & 62.00 & \textbf{62.88} & 61.54 & 56.10 & 56.22 & 56.60 & 56.85 & 5.11 $\uparrow$  \\ \hline \hline
\textbf{BLURB} & 55.20 & 56.27 & 57.41 & \textbf{58.03} & 57.12 & 53.91 & 54.26 & 53.83 & 53.82 & 2.83$\uparrow$  \\ \hline \hline
\end{tabular}
\\[6pt]
{\small \textbf{Note:} The ‘difference over SOTA’  indicate the absolute change ($\uparrow$ for increase and $\downarrow$ for decrease) in metric performance over SOTA.   \textbf{Bold} has the best results. We report the SOTA model results on various datasets as follows: (i) JNLPBA, BC2GM, ChemProt, and GAD from Yuan et al.\cite{yuan2021improving} (KeBioLM), (ii) DDI ad BIOSSES are from Gu et al.~\cite{gu2020domain} (PubMedBERT),   (iii) Share/Clefe, i2b2, MedSTS, MedNLI and HoC  from Peng et al.~\cite{peng2019transfer} (BLUE), (iv) BC5CDR (disease), BC5CDR (chemical), NCBI (Disease), S800, Euadr, BioASQ 4b, BioASQ 5b, and BioASQ 6b, from Lee et al.~\cite{lee2019biobert} (BioBERT), and (v) LINNAEUS from Giorgi and Bader~\cite{giorgi2018transfer}.  The Biomedical Language Understanding \& Reasoning Benchmark (BLURB) is an average score among all tasks used in previous studies~\cite{gu2020domain,yuan2021improving}. 
}
\end{table*}

\section*{Results and Discussion} \label{aaamodel}

\begin{itemize}

\item \textbf{Comparison with SOTA biomedical LMs:} Table~\ref{results} summarizes the results\footnote{Refer to Table~\ref{versions} for more details of BioALBERT size and training corpus and Table~\ref{stats} for the evaluation metric used in each dataset.} for all the BioALBERT variants in comparison to the baselines~\footnote{Baseline results were acquired from the respective original publication.}. We observe that the performance of BioALBERT\footnote{Here, we discuss the best model of BioALBERT Out of 8 versions of BioALBERT.} is higher than SOTA models on 17 datasets out of 20 datasets and in 5 out of the 6 tasks. Overall, a large version of ALBERT trained on PubMed abstract achieved the best results among all the tasks.


For NER, BioALBERT was significantly higher compared to SOTA methods on all 8 datasets (ranging from 4.61\% to 23.71\%) and outperformed the SOTA models by 11.09\% in terms of micro averaged F1-score (BLURB score). For, Share/Clefe dataset, BioALBERT increased the performance by 19.44\%, 10.63\% for BC5CDR-disease, 4.61\% for BC5CDR-chemical, 4.74\% for JNLPBA, 6.19\% for Linnaeus, 7.47\% for NCBI-disease, 23.71\% and 12.25\% for S800 and BC2GM datasets, respectively.


For RE, BioALBERT outperformed SOTA methods on 3 out of 5 datasets by 1.69\%, 0.82\%, and 0.46\% on DDI, ChemProt and i2b2 datasets, respectively.  On average (micro), BioALBERT obtained a higher F1-score (BLURB score) of 0.80\% than the SOTA LMs. For Euadr and GAD performance of BioALBERT slightly drops because the splits of data used are different. We used an official split of the data provided by authors, whereas the SOTA method reported 10-fold cross-validation results; typically, having more folds increase results.  

For STS, BioALBERT achieved higher performance on both datasets by a 1.05\% increase in average Pearson score (BLURB score) as compared to SOTA models. In particular, BioALBERT achieved improvements of 0.50\% for BIOSSES and 0.90\% for MedSTS.



Similarly, for document classification, BioALBERT slightly increase the performance by 0.62\% for the HoC dataset and the inference task (MedNLI dataset), the performance of BioALBERT drops slightly, and we attribute this to the average length of the sentence being smaller compared to others.


For QA, BioALBERT achieved higher performance on all 3 datasets and increased average accuracy (lenient) score (BLURB score) by 2.83\% compared to SOTA models. In particular, BioALBERT improves the performance by 1.08\% for BioASQ 4b, 2.31\% for BioASQ 5b and 5.11\% for BioASQ 6b QA datasets respectively as compared to SOTA.


We note that the performance of ALBERT (both base and large), when pre-trained on MIMIC-III, in addition to PubMed and combination of PubMed and PMC, drops as compared to the same pre-trained ALBERT without MIMIC-III, especially in RE, STS and QA tasks. Since MIMIC consists of notes from the ICU of Beth Israel Deaconess Medical Center (BIDMC) only, the data size was relatively smaller than PubMed and PMC, and therefore we suggest that the BioALBERT performance was poor when compared to models trained on PubMed only or PubMed and PMC. BioALBERT (large), trained on PubMed with dup-factor as five, performed better.


\begin{table}[!t]
\centering
\small
\caption{Comparison of run-time (in days) statistics of BioALBERT v/s BioBERT. Refer to Table~\ref{versions} for more details of BioALBERT size. BioBERT$_{Base1}$ and BioBERT$_{Base2}$ refers to BioBERT trained on PubMed and PubMed+PMC respectively}
\label{runstats}
\begin{tabular}{cc}
\hline \hline
Model & Training time \\ \hline \hline
BioBERT$_{Base1}$     & 23.00 \\
BioBERT$_{Base2}$   & 10.00 \\ \hline \hline
BioALBERT$_{Base1}$   & 3.00 \\
BioALBERT$_{Base2}$  & 4.08 \\
BioALBERT$_{Large1}$ & 2.83 \\
BioALBERT$_{Large2}$   & 3.88 \\
BioALBERT$_{Base3}$   & 4.02 \\
BioALBERT$_{Base4}$   & 4.45 \\
BioALBERT$_{Large3}$  & 4.62 \\
BioALBERT$_{Large4}$  & 4.67 \\ \hline \hline
\end{tabular}
\end{table}

\begin{table}[!b]
\centering
\small
\caption{Prediction samples from ALBERT and BioALBERT. Bold entities are better recognised by BioALBERT}
\label{pred}
\hspace*{-1.0cm}
\begin{tabular}{ccl}
\hline \hline
Dataset & Model & \multicolumn{1}{c}{Sample} \\ \hline \hline
\multirow{2}{*}{JNLPBA} & ALBERT & Number of glucocoticoid receptors in lymphocytes and their sensitivity to… \\
 & BioALBERT & Number of glucocoticoid \textbf{receptors} in \textbf{lymphocytes} and their   sensitivity to… \\ \hline
\multirow{2}{*}{Share/Clefe} & ALBERT & The mitral valve leaflets are mildly thickened . There is mild mitral   annular calcification .TRICUSPID VALVE… \\
 & BioALBERT & The mitral valve leaflets   are mildly \textbf{thickened} .   There is mild \textbf{mitral annular calcification} .TRICUSPID VALVE… \\ \hline
\multirow{2}{*}{HoC} & ALBERT & In contrast , 15 Gy increased the expression of p27 in radiosensitive   tumors and reduced it in radioresistant tumors. \\
 & BioALBERT & In contrast , 15 Gy increased the expression of p27 in radiosensitive \textbf{tumors} and reduced it in \textbf{radioresistant tumors}. \\ \hline \hline
\end{tabular}
\end{table}

We compared pre-training run-time statistics of BioALBERT with BioBERT with all variants of BioALBERT outperforming BioBERT. The difference in performance is significant, identifying BioALBERT as a robust and practical model. BioBERT$_{Base}$ trained on PubMed took 10 days, and BioBERT$_{Base}$ trained on PubMed and PMC took 23 days, whereas all models of BioALBERT took less than 5 days for training an equal number of steps. The run time statistics of both pre-trained models are given in Table~\ref{runstats}. \\


\begin{figure}[!htpb]
\centering
\includegraphics[width=1\linewidth]{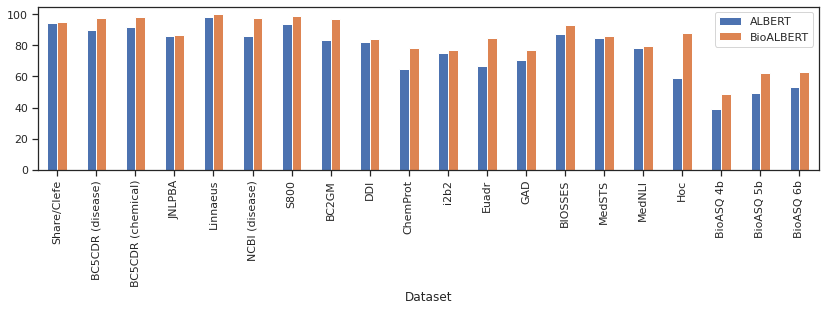}
\caption{Comparison of BioALBERT v/s ALBERT}
\label{bio}
\end{figure} 

\item \textbf{Comparison with SOTA general LM (ALBERT):} We compared the performance of ALBERT trained on general corpora to BioALBERT with the results shown in Figure~\ref{bio}. BioALBERT consistently achieved higher performance on all 6 tasks (20 out of 20 datasets) as compared to ALBERT. Further, as shown in Table~\ref{pred}, we sampled predictions from ALBERT and BioALBERT to see the effect of pre-training on downstream tasks. BioALBERT can better recognise the biomedical entities compared to ALBERT both in NER (JNLPBA and Share/Clefe) and document classification (HoC) datasets.

\end{itemize}

\section*{Conclusion} \label{aaamodel}


We present BioALBERT, the first adaptation of ALBERT trained on both biomedical text and clinical data. Our experiments show that  training general domain language models on domain-specific corpora leads to an improvement in performance across a range of biomedical BioNLP tasks. BioALBERT outperforms previous state-of-the-art  models on 5 out of 6 benchmark tasks  and on 17 out of 20 benchmark datasets. Our expectation is that the release of the BioALBERT models and data will support the development of new applications built from biomedical NLP tasks.

\bibliography{sample}




\section*{Code Availability}

Pre-trained weights of BioALBERT models and reproduced results of the benchmarks presented in this paper is 
at \url{https://github.com/usmaann/BioALBERT.}



\section*{Competing interests} 

The authors declare no competing interests.

\end{document}